# Adaptive Canonical Correlation Analysis Based On Matrix Manifolds

**Florian Yger** FLORIAN.YGER@LITISLAB.EU
**Maxime Berar** MAXIME.BERAR@LITISLAB.EU
**Gilles Gasso** GILLES.GASSO@INSA-ROUEN.FR
**Alain Rakotomamonjy** ALAIN.RAKOTO@INSA-ROUEN.FR
LITIS/Université de Rouen-INSA de Rouen, Avenue de l'Université, St-Etienne-du-Rouvray, 76800 France

## Abstract

In this paper, we formulate the Canonical Correlation Analysis (CCA) problem on matrix manifolds. This framework provides a natural way for dealing with matrix constraints and tools for building efficient algorithms even in an adaptive setting. Finally, an adaptive CCA algorithm is proposed and applied to a change detection problem in EEG signals.

## 1. Introduction

Canonical Correlation Analysis (CCA) is a well-known dimensionality reduction method. Given two views (or representations) of the same set of objects, it aims at finding projections for each representation such that their correlation is maximized in the projection space. As every popular method in machine learning, since its first formulation (Hotelling, 1936) CCA has been extended to a kernel version (Lai & Fyfe, 2000; Akaho, 2001), to online and recursive versions (Vía et al., 2007) and quite recently to a sparse version (Hardoon & Shawe-Taylor, 2011).

CCA is usually formulated as the Generalized Singular Value Decomposition (Generalized SVD) of the cross-covariance matrix (Sun et al., 2009). Besides, it aims at finding projections that are orthogonal with respect to the auto-covariance matrices of each view. As CCA belongs to the class of Latent Variables methods, it shares close connections with those methods. Indeed, according to Rosipal & Krämer (2006); Sun et al. (2009), CCA is a generalization of Orthonormalized Partial Least Squares.

A wide variety of applications from multimodal deep learning (Ngiam et al., 2011) to Brain Computer Interface (BCI) (Hakvoort et al., 2011) used CCA for feature extraction. However, those applications involve either non-stationary or big datasets. In such contexts, an adaptive algorithm for solving CCA would be useful. Incremental algorithms have been proposed for CCA and are based on a Recursive Least Squares algorithm (Vía et al., 2007), but they solve a sequence of rank-one CCA problems and cope with the orthogonality constraints using a deflation scheme.

Differential geometry of matrix manifolds provides an elegant way for dealing with matrix constraints. This framework has proved competitive for matrix constrained optimization problems (Meyer et al., 2011) and more specifically for eigenproblems (Absil et al., 2008). In this contribution, we cast the CCA problem as an optimization problem on matrix manifolds, and solve it using classical gradient algorithms on less studied manifolds. As an application, our adaptive CCA is used to track principal correlations subspaces of EEG signals over time, in order to detect the exact time step when abrupt changes occur in those subspaces. In our BCI context, such a change means that the patient switches from a mental task to another. For a real-time BCI system, a fast, reliable, low rank and adaptive algorithm is needed to handle noisy and non-stationary signals (Millan et al., 2004). Keeping those needs in mind, we developed an adaptive version of CCA based on matrix manifolds.

The next section of this paper is devoted to the CCA algorithm and to its adaptive formulation. Then, we introduce the basic tools of Riemannian geometry used in our algorithm and we describe the steps of our CCA algorithm. Finally, before discussing some extensions of this work, we present numerical results on a toy dataset and on BCI data.

## 2. Canonical Correlation Analysis

### 2.1. Background

Assume given two full-rank data matrices $X \in \mathbb{R}^{n\times T}$ and $Y \in \mathbb{R}^{m\times T}$. The matrices $C_{xy} = X^\top Y$, $C_x = X^\top X$, $C_y = Y^\top Y$ are estimates of the cross-covariance and auto-covariance matrices. We assume that both views on the data are centered.

Usually, CCA is formulated in terms of Rayleigh Quotient (Hardoon et al., 2004; Bie et al., 2005) involving the covariance matrices of both views $C_x$ and $C_y$ and between the views $C_{xy}$:

$$\max_{u\in\mathbb{R}^n, v\in\mathbb{R}^m} \frac{u^\top C_{xy} v}{\sqrt{u^\top C_x u}\sqrt{v^\top C_y v}}. \quad (1)$$

By extending the Rayleigh Quotient using the trace operator, multiple projections can be obtained simultaneously by solving the following fixed rank $p$ optimization problem (Sun et al., 2009)

$$\begin{cases} \max\limits_{\substack{U\in\mathbb{R}^{n\times p}\\ V\in\mathbb{R}^{m\times p}}} & tr\left(U^\top C_{xy} V\right) \\ \text{s.t. } U^\top C_x U = \mathbb{I}_p, & V^\top C_y V = \mathbb{I}_p. \end{cases} \quad (2)$$

Several alternative formulations exist and involve Generalized Eigenvalue Decompositions of symmetric definite positive matrix pencils (Golub & Van Loan, 1996):

$$\begin{cases} C_{xy} C_y^{-1} C_{xy}^\top U = C_x U \Lambda^2, \\ C_{xy}^\top C_x^{-1} C_{xy} V = C_y V \Lambda^2. \end{cases}$$

However, if the data are non-stationary, the previous formulations are hard to update, making them unsuitable for online or adaptive applications.

Concatenating the two views, the following eigenproblem is often used for the CCA problem (Bie et al., 2005; Vía et al., 2007):

$$\frac{1}{2}\begin{bmatrix} 0 & C_{xy} \\ C_{xy}^\top & 0 \end{bmatrix}\begin{bmatrix} u \\ v \end{bmatrix} = \lambda \begin{bmatrix} C_x & 0 \\ 0 & C_y \end{bmatrix}\begin{bmatrix} u \\ v \end{bmatrix}.$$

This formulation, true for rank-one eigenproblem, cannot be extended to a higher rank without a costly and numerically unstable deflation procedure (unsuitable for an adaptive formulation). Indeed, for $p > 1$ the constraint $U^\top C_x U + V^\top C_y V = \mathbb{I}_p$ implied by the formulation is not equivalent to the constraints of (2).

As the trace operator is invariant to multiplication of both $U$ and $V$ by elements of $\mathcal{O}(p)$, the group of orthonormal matrices in $\mathbb{R}^{p\times p}$, we must modify Problem (2) to enforce uniqueness of the solution. Introducing the matrix $N$ as a diagonal matrix in $\mathbb{R}^{p\times p}$ with strictly decreasing positive elements (Absil et al., 2008, p.11), we propose to replace the objective function of Problem (2) by the Brockett cost function $tr(U^\top C_{xy} V N)$. Hence the obtained solution will correspond to the solutions of the previous eigenproblems.

### 2.2. Adaptive formulation

At each time step $t$, new samples $x_t$ in $\mathbb{R}^n$ and $y_t$ in $\mathbb{R}^m$ are acquired and all covariance matrices are updated using a forgetting factor $0 < \beta < 1$:

$$\begin{aligned} C_x^t &= \beta C_x^{t-1} + x_t x_t^\top, \quad C_y^t = \beta C_y^{t-1} + y_t y_t^\top \\ C_{xy}^t &= \beta C_{xy}^{t-1} + x_t y_t^\top. \end{aligned}$$

This forgetting update is a common tool in subspace tracking (dos Santos Teixeira & Milidiú, 2010), assimilable to estimation over an exponential window. The case $\beta = 1$ leads to an incremental problem.

The adaptive CCA problem is formulated as

$$\begin{cases} \max\limits_{\substack{U\in\mathbb{R}^{n\times p}\\ V\in\mathbb{R}^{m\times p}}} & tr\left(U^\top\left(\beta C_{xy}^{t-1} + x_t y_t^\top\right) V N\right) \\ \text{s.t. } U^\top\left(\beta C_x^{t-1} + x_t x_t^\top\right) U = \mathbb{I}_p, \\ \quad\;\; V^\top\left(\beta C_y^{t-1} + y_t y_t^\top\right) V = \mathbb{I}_p \end{cases} \quad (3)$$

In our adaptive setting, knowing the solution $(U_{t-1}, V_{t-1})$, we wish to update the solution at a cheap cost. However, we need our solution $(U_t, V_t)$ to satisfy the $C_x^t$- and $C_y^t$-orthogonality conditions. In the sequel, we present a gradient algorithm, but, in order to be able to apply this algorithm, we need the initial point $(U_{t-1}, V_{t-1})$ to satisfy the updated constraints. Indeed, if we have $U_{t-1}^\top C_x^{t-1} U_{t-1} = \mathbb{I}_p$ the constraint $U_{t-1}^\top C_x^t U_{t-1} = \mathbb{I}_p$ may not be satisfied. Hence, our first task is to find a feasible starting point referred to as $(U_{t'}, V_{t'})$. We call this the metric update subproblem. From then, our second task is to maximize the cost function.

#### 2.2.1. Metric update subproblem

The metric update consists in finding a subspace satisfying the new orthogonality constraints while conserving the span of the previous subspace. The product by matrices in the general linear group $GL_p$ (the set of all invertible $p\times p$ matrices) conserves the span. The metric problem then aims at finding matrices in $GL_p$ such that:

$$\begin{cases} \max\limits_{\substack{O_u\in GL_p\\ O_v\in GL_p}} & tr\left(O_u^\top U_{t-1}^\top C_{xy}^{t-1} V_{t-1} O_v N\right) \\ \text{s.t. } O_u^\top\left(\beta\mathbb{I}_p + U_{t-1}^\top x_t x_t^\top U_{t-1}\right) O_u = \mathbb{I}_p, \\ \quad\;\; O_v^\top\left(\beta\mathbb{I}_p + V_{t-1}^\top y_t y_t^\top V_{t-1}\right) O_v = \mathbb{I}_p \end{cases}$$

Let $z_x = U_{t-1}^\top x_t$ and $z_y = V_{t-1}^\top y_t$ be the compressed samples, $L_{t-1} = U_{t-1}^\top C_{xy}^{t-1} V_{t-1}$ be the compressed cross-covariance and $G_x = \beta \mathbb{I}_p + z_x z_x^\top$ and $G_y = \beta \mathbb{I}_p + z_y z_y^\top$ be the compressed auto-covariance matrices.

The only elements of $GL_p$ respecting the constraints can be written as elements of two instances of the Generalized orthogonal group defined for a given matrix $G \succ 0$ (positive definite matrix $G$) as:

$$\mathcal{O}_G(p) = \{X \in GL(p) : \; X^\top G X = \mathbb{I}_p\}.$$

The metric update problem is now a full SVD problem on the Generalized orthogonal groups:

$$\begin{cases} \max\limits_{O_u, O_v} tr\left(O_u^\top L_{t-1} O_v N\right) \\ \text{s.t. } O_u \in \mathcal{O}_{G_x}(p),\; O_v \in \mathcal{O}_{G_y}(p) \end{cases} \tag{4}$$

From the solution of (4), the new subspace matrices are obtained as

$$U_{t'} = U_{t-1} O_u, \quad V_{t'} = V_{t-1} O_v,$$

along with a new compressed covariance matrix

$$L_{t'} = U_{t'}^\top C_{xy}^{t-1} V_{t'} = O_u^\top L_{t-1} O_v.$$

### 2.2.2. Cost function update subproblem

This subproblem consists in finding new subspaces maximizing the cost while satisfying the updated metric constraints, (eventually) resulting in a change of span and it is equivalent to Problem 3:

$$\begin{cases} \max\limits_{\substack{U \in \mathbb{R}^{n \times p} \\ V \in \mathbb{R}^{m \times p}}} \; tr\left(U^\top \left(\beta C_{xy}^{t-1} + x_t y_t^\top\right) V N\right) \\ \qquad \text{s.t.} \quad U^\top C_x^t U = \mathbb{I}_p, \\ \qquad\qquad V^\top C_y^t V = \mathbb{I}_p \end{cases}$$

At the end of the cost function phase, the new subspace matrices $U_t$ and $V_t$ are obtained directly as solutions of the subproblem, and so is the new compressed covariance matrix $L_t = U_t^\top C_{xy}^t V_t$.

In each subproblem, the matrix constraints define Riemannian matrix manifolds. Hence, each subproblem corresponds to the maximization of a Brockett cost function on matrix manifolds. In the case of the metric update, the manifold is a product of two Generalized orthogonal groups and in the case of the cost function update, the manifold is a product of Generalized Stiefel manifolds.

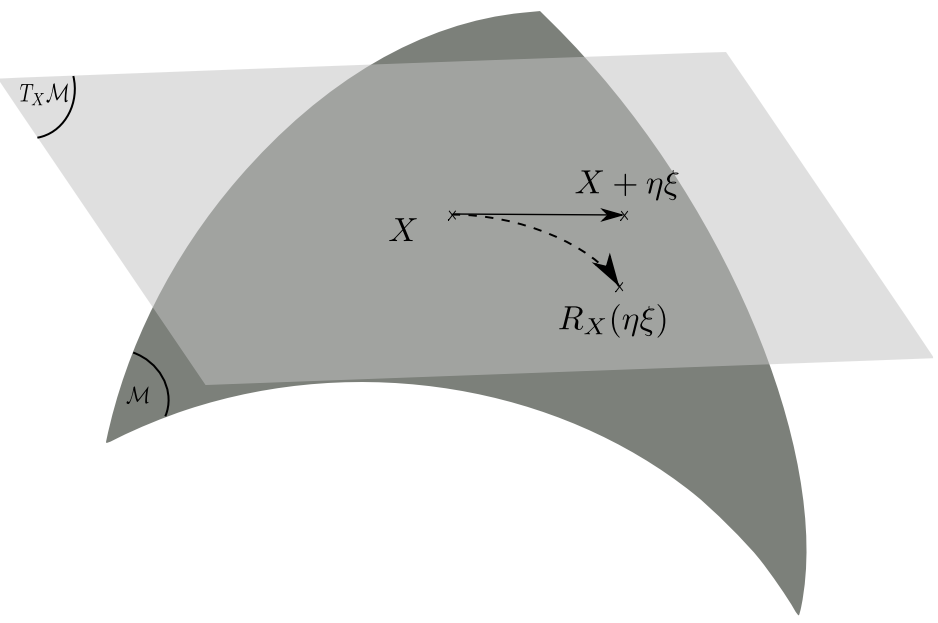

*Figure 1.* Insights on Riemannian geometry : the link between a manifold $\mathcal{M}$, its tangent space $T_X\mathcal{M}$ at a point $X$ and the retraction $R_X$ that locally maps a displacement $\eta\xi$ in $T_X\mathcal{M}$ to $\mathcal{M}$.

## 3. Manifold framework

Optimization on Riemannian manifold is currently a very active research field in machine learning community (Meyer et al., 2011) and more broadly in the numerical optimization discipline (Absil et al., 2008). In a complete Riemannian view, we should express our optimization problem as a search along a geodesic curve in the manifold. Such a search being in practice intractable, we chose to approximate it by a search along another smooth curve on the manifold. This smooth curve is defined by a function that transforms any displacement in the tangent space to a point on the manifold. Such a function (that also obeys to some technical conditions, see (Absil et al., 2008)) is called a retraction. Figure 1 depicts a Riemannian manifold $\mathcal{M}$ and a tangent space at a point $X$ on this manifold. The tangent space is a vector space that locally approximates the manifold. Then the retraction is a mapping that locally transforms a search in the manifold into a search in the tangent space.

In this section, we briefly recall our Tangent space and retraction formulae. The Stiefel manifold $St(p, n)$ is defined as

$$St(p, n) = \{X \in \mathbb{R}^{n \times p} : \; X^\top X = \mathbb{I}_p\}.$$

The orthogonal group $\mathcal{O}(p)$ is the Stiefel manifold $St(q, p)$ with $q = p$: $\mathcal{O}(p) = \{X \in \mathbb{R}^{p \times p} : \; X^\top X = \mathbb{I}_p\}$. For a given matrix $G \succ 0$, we define the Generalized Stiefel manifold as:

$$St_G(p, n) = \{X \in \mathbb{R}^{n \times p} : \; X^\top G X = \mathbb{I}_p\}.$$

This manifold contains the $p$-dimensional $G$-orthonormal subspaces of $\mathbb{R}^{n \times n}$. The Generalized orthogonal group $\mathcal{O}_G(p)$ is the Generalized Stiefel manifold $St(q, p)$ with $q = p$:

$$\mathcal{O}_G(p) = \{X \in \mathbb{R}^{p \times p} : \; X^\top G X = \mathbb{I}_p\}.$$

The former CCA optimization problems can be now seen as optimization problems on matrix manifolds $St_{C_x}(p,n)$ and $St_{C_y}(p,m)$:

$$\left\{ \max_{\substack{U \in St_{C_x}(p,n)\\ V \in St_{C_y}(p,m)}} tr\left(U^\top C_{xy} V N\right) \right. \tag{5}$$

and on manifolds $\mathcal{O}_{G_x}(p)$ and $\mathcal{O}_{G_y}(p)$ as in Eq. 4.

### 3.1. Tangent space

The definitions of the tangent space and of the retraction can be found in (Absil et al., 2008) for the Stiefel manifold and for the orthogonal group. Deriving the same calculus for the generalized case, we obtain the following definitions for the tangent space at a point $X$ on a Generalized Stiefel manifold:

$$T_X St_G(p,n) = \{Z \in \mathbb{R}^{n\times p} : \; X^\top G Z + Z^\top G X = 0\}.$$

An alternative characterization of $T_X St_G(p,n)$ decomposes any element of $T_X St_G(p,n)$ as the sum of a term $G$-orthogonal to $X$, written $X_\perp K$, and of the product of $X$ with a skew-symmetric matrix $\Omega$ in $\mathcal{S}_{skew}(p)$:

$$T_X St_G = \{X\Omega + X_\perp K : \; \Omega^\top = -\Omega, K \in \mathbb{R}^{(n-p)\times p}\}.$$

where $\mathcal{S}_{skew}(p)$ denotes the set of all skew-symmetric $p\times p$ matrices. In the case of the Generalized orthogonal group, the alternative characterization of the tangent tangent space is simplified:

$$T_X \mathcal{O}_G(p) = \{X\Omega : \; \Omega^\top = -\Omega\}.$$

### 3.2. Retraction

One example of retraction for the (Generalized) Stiefel manifold is the polar retraction:

$$\forall \xi \in T_X St_G(n,p) \text{ and } X \in St_G(n,p) \;:$$
$$R_X(\zeta\xi) = (X + \zeta\xi)(\mathbb{I}_p + \zeta^2 \xi^\top G \xi)^{-\frac{1}{2}}$$

There exists several other retractions that can be applied for (Generalized) Stiefel manifold. Among them, the retraction based on QR-decomposition (adapted to the metric) could also be applied.

Until now, we derived equations and formulae characterizing tangent space and retraction on $St_G(n,p)$ and $\mathcal{O}_G(p)$. However, in our optimization problems, the manifold of interest is a product manifold of two generalized Stiefel manifolds.
In (Ma et al., 2001), the authors proved that the geodesics in the product manifold are the products of the geodesics in the factor manifolds. This helpful property enables us to compute the gradients and the retractions on each of the factor manifolds separately.

## 4. Gradient ascend algorithm

This section presents a manifold gradient algorithm adapted to product of Generalized Stiefel manifolds and product of Generalized orthogonal groups. First, we present the formulae for the cost function subproblem. The Generalized orthogonal group being a particular case of Generalized Stiefel manifold, we shortened the description of its gradient. Finally, we sum up the approach in Algorithm 1.

### 4.1. Gradient ascend on Generalized Stiefel manifold

At time $t'$, after the metric update phase, for each subspace, we can compute compressed samples and residuals:

$$\begin{aligned} z_x &= U_{t'}^\top x_t, & f_x &= (C_x^t)^{-1} x - U_{t'} z_x, \\ z_y &= V_{t'}^\top y_t, & f_y &= (C_y^t)^{-1} x - V_{t'} z_y, \end{aligned} \tag{6}$$

using Sherman-Morrison-Woodbury formula to get efficiently the inverse of matrices. For instance, we have

$$(C_x^t)^{-1} = \beta^{-1}(C_x^t)^{-1} - \beta^{-1}\frac{(C_x^t)^{-1} x_t x_t^\top (C_x^t)^{-1}}{\beta + x_t^\top (C_x^t)^{-1} x_t}.$$

The gradient of the Brockett cost function on the Generalized Stiefel manifold $St_{C_x}$ at $U$ is defined as :

$$\xi_U = (C_x)^{-1} C_{xy} V N - \frac{1}{2} U L N - \frac{1}{2} U N L^\top,$$

with $L = U^\top C_{xy} V$. Note that the Brockett matrix $N$ insures that if the gradient is null then the compressed covariance matrix $L$ is diagonal and $U$ is solution of the eigenproblem. In the adaptive case, the gradient becomes

$$\xi_U = f_x z_y^\top N + \frac{1}{2} U_{t'} \left( z_x z_y^\top N - N z_y z_x^\top \right). \tag{7}$$

Defined as the sum of a $C_x^t$-orthogonal element and of a skew-symmetric element, $\xi_U$ belongs to the tangent space $T_U St(p,n)$ and thus a retraction can be used.

Choosing a polar retraction, the updated matrix $U_{t+1}$ for a step length $\zeta_U$ in $\mathbb{R}$ is :

$$U_{t+1} = (U_{t'} + \zeta_U \xi_U)\left(I_p + \zeta_U^2 \xi_U^\top C_x \xi_U\right)^{-1/2}$$

The right term of the retraction product includes two rank-one matrices orthogonal to each other, multiple of the (orthogonal) projectors on $N z_y$ and $\bar{z}_x$, where $\bar{z}_x \triangleq z_x - \frac{z_x^\top N z_y}{z_y^\top N^2 z_y} N z_y$ with $\bar{z}_x^\top z_y = 0$. After some algebra, the updated matrix can be computed efficiently as:

$$U_{t+1} = U_{t'} - \rho_x U_{t'} \frac{N z_y z_y^\top N}{z_y^\top N^2 z_y} - \bar{\rho}_x U_{t'} \frac{\bar{z}_x \bar{z}_x^\top}{\bar{z}_x^\top \bar{z}_x} \quad (8)$$

$$+ \zeta_U (1-\rho_x) f_x z_y^\top N + \frac{\zeta_U}{2}(1-\rho_x) U_{t'} \bar{z}_x z_y^\top N$$

$$- \frac{\zeta_U}{2}(1-\bar{\rho}_x) U_{t'} N z_y \bar{z}_x^\top .$$

with the coefficients $\rho_x$ and $\bar{\rho}_x$ defined as:

$$\rho_x = 1 - \sqrt{\frac{1}{1+\alpha_x}},\ \bar{\rho}_x = 1 - \sqrt{\frac{1}{1+\bar{\alpha}_x}},$$

where $\alpha_x$, $\bar{\alpha}_x$ are positive coefficients taking into account the norm of the residuals, the compressed samples and the squared step length:

$$\alpha_x = (f_x^\top C_x f_x + \frac{\bar{z}_x^\top \bar{z}_x}{4})(z_y^\top N^2 z_y)\zeta_U^2,$$

$$\bar{\alpha}_x = \frac{\bar{z}_x^\top \bar{z}_x}{4}(z_y^\top N^2 z_y)\zeta_U^2.$$

The same formula holds for the gradient of the right subspace exchanging the respective roles of each view. The step lengths $\zeta_U$ and $\zeta_V$ are jointly determined using standard line-search techniques.

### 4.2. Gradient ascend on Generalized Orthogonal Group

Let $G_x = \beta\mathbb{I}_p + z_x z_x^\top$ and $G_y = \beta\mathbb{I}_p + z_y z_y^\top$ be the compressed auto-covariance matrices. Initial points on the manifold are computed using the following formulae:

$$O_u^t = (G_x)^{-1/2} = \beta^{-1/2}\left(\mathbb{I}_p - \tilde{\rho}_x \frac{z_x z_x^\top}{z_x^\top z_x}\right), \quad (9)$$

$$\tilde{\rho}_x = 1 - \sqrt{\frac{1}{1+\tilde{\alpha}_x}},\ \tilde{\alpha}_x = \frac{z_x^\top z_x}{\beta(1+z_x^\top z_x)},$$

$$O_v^t = \beta^{-1/2}\left(\mathbb{I}_p - \tilde{\rho}_y \frac{z_y z_y^\top}{z_y^\top z_y}\right), \quad (10)$$

$$\tilde{\rho}_y = 1 - \sqrt{\frac{1}{1+\tilde{\alpha}_y}},\ \tilde{\alpha}_y = \frac{z_y^\top z_y}{\beta(1+z_y^\top z_y)},$$

and at these points, the gradient is written:

$$\xi_{O_u} = (G_x)^{-1} L_t O_v^t N - \frac{1}{2} O_u^t L_t N - \frac{1}{2} O_u^t N L_t^\top, \quad (11)$$

A polar retraction being too costly in this case, we apply an oblique version of $QR$ decomposition, where the projections are made according to $G_x$. Again, the same formula holds for the gradient of the right subspace exchanging the respective roles of each view.

**Algorithm 1** Adaptive CCA algorithm

**input:** subspace matrices $(U,V)$, forgetting factor $\beta$, covariance matrices $C_x$,$C_y$,$C_{xy}$, sample $(x,y)$
**output:** updated subspace matrices $(U,V)$ updated covariance matrices $C_x$,$C_y$,$C_{xy}$

compute compressed samples $z_x$,$z_y$(Eq. 6)
compute initial matrix $O_u$,$O_v$(Eq.9-10)
perform Gradient ascend (Eq. 11)
update $U,V$
update auto-covariance matrices $C_x$,$C_y$
compute $z_x$,$z_y$ and residuals $f_x$,$f_y$ (Eq. 6)
perform Gradient ascend (Eq. 7)
update $U,V$ (Eq. 8)
update cross-covariance matrix $C_{xy}$

## 5. Numerical results

In the begining of this paper, in preamble to the explanation on CCA, we assumed both views of the data to be centered. This theoretical requirement being however rarely met in practice, we used the following formula (dos Santos Teixeira & Milidiú, 2010) in order to estimate the mean of a given view $w$(used to center the current sample): $\mu_w^t = \frac{t-1}{t}\beta\mu_w^{t-1} + \frac{1}{t}(w_t + \mu_w^{t-1})$.

### 5.1. Toy dataset

We compare the performances of our algorithm to those of the adaptive RLS-CCA algorithm (Vía et al., 2007) with a corrected orthogonalization scheme. The solutions given by both methods are compared to the exact solution given by a batch algorithm (solving the Generalized SVD at every step). The evaluation criteria are the orthonormality errors w.r.t. the metric on both views

$$e_o^x(t) = \|U_t^\top C_x^t U_t - \mathbb{I}_p\|_F^2,\ e_o^y(t) = \|V_t^\top C_y^t V_t - \mathbb{I}_p\|_F^2,$$

the distance between the oblique projectors associated to each subspace normalized by the rank of the problem:

$$e_a^x(t) = \frac{\|C_x^t U_t U_t^\top - C_x^t U_t^b U_t^{b\ \top}\|_F^2}{2p},$$

$$e_a^y(t) = \frac{\|C_y^t V_t V_t^\top - C_x^t V_t^b V_t^{b\ \top}\|_F^2}{2p},$$

and the ratio $e_c$ between the cost obtained using the competing algorithms and the cost of the exact method.

Our simulation protocol generates samples $x$ of size $n = 36$ and $y$ of $m = 34$ living in two different full-rank subspaces. We are interested in tracking the first $p = 30$ principal directions of the correlated subspaces with a forgetting factor $\beta = 0.99$ during 2000 time steps. The results are averaged over 50 trials.

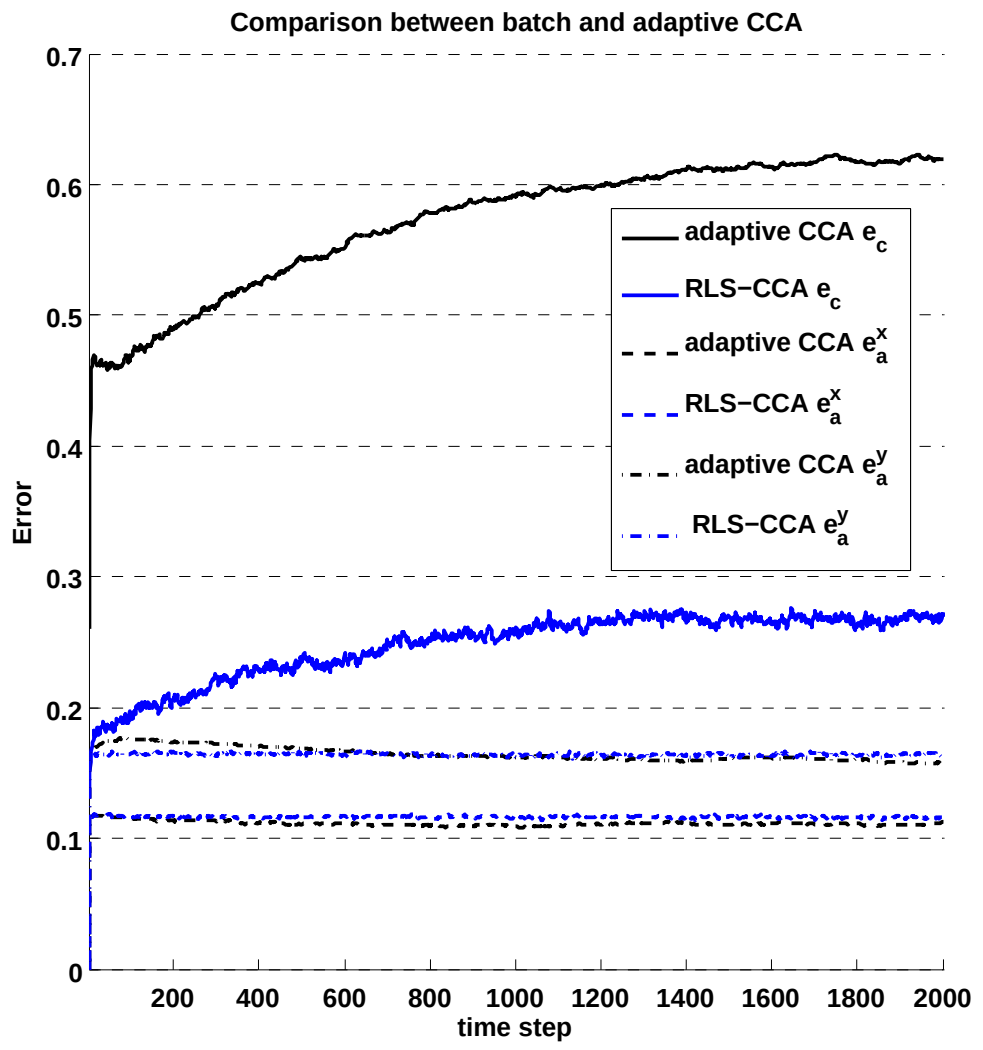

*Figure 2.* Cost and accuracy distance to the batch solution for the RLS-CCA and our adaptive CCA

Figure 2 shows the cost and accuracy errors with random feasible initializations. The error on the obtained cost stabilizes at 62% of the batch cost for our algorithm, whereas the LRR-CCA stabilizes at 27% of the batch cost algorithm. In both cases, the accuracy errors are in the same range and the orthogonality errors are acceptable ($\sim 10^{-13}$). As our adaptive algorithm only performs one gradient step for every sample, comparison with the updated batch is far from being optimal. But it achieves correct results, compatible with our change detection application for BCI.

### 5.2. BCI Competition III dataset V

We tested our method on the dataset V of the BCI Competition III (Millan et al., 2004; Blankertz et al., 2004). It contains data from three subjects during 4 non-feedback sessions. The subjects were asked to perform a mental task during about 15 seconds and then switch to a randomly chosen task. There are three tasks : imagination of repetitive left hand movements, or imagination of repetitive right hand movements or generation of words beginning with the same random letter. The goal is to predict the current mental task of the subject every 0.5 seconds.

In this study, we apply our adaptive CCA algorithm for change detection and we focus on detecting the change in mental task of the subjects. Some precomputed features were provided for the competition. Those 96 features consists of power spectral density (PSD) in the band $[8 - 30]$ Hz of the 8 centro-parietal channels, spatially filtered by a surface Laplacian. We will calculate the correlation between the features of left and right electrodes, respectively $x_t$ and $y_t$, letting unused the features of the central electrodes.

The three mental tasks performed by the subjects involve different regions of the brain. Hence, the correlation of the signals measured over those regions will be different for two different mental tasks. So, the correlation matrix between the right and left electrodes should give us information about the current mental state of the brain. By tracking the changes in the principal subspaces of this matrix, we should be able to detect if the subject switched from one mental task to another.

The cross-covariance and auto-covariance matrices were initialized on the 100 first samples ($\approx$ 6s) of each session. This choice is a compromise between an accurate estimation of the covariance matrices and the need for the criterion to be stable before the first change. The forgetting factor $\beta$ was set to 0.98 such that it avoids numerical problem and gives a sufficient adaptability to the algorithm. Finally, the rank of the decomposition was set to $p = 4$ subspaces. This choice is a trade-off between summing up most of the views correlation and keeping a low rank representation for a fast optimization.

In (dos Santos Teixeira & Milidiú, 2010), a criterion involving the reconstruction error on the current sample has been used for anomaly detection. We adapted this criterion to our context and used it in order to detect the change in mental task of the subjects. Let $n_x = n_y = 36$, the number of features in both views $x_t$ and $y_t$. For a given sample, we calculate the residuals $r_x^t$ and $r_y^t$ of the projection of the current views on the previous subspaces. $r_x^t = ((C_x^{t-1})^{-1} - U_{t-1}U_{t-1}^\top)x_t$, $r_y^t = ((C_y^{t-1})^{-1} - V_{t-1}V_{t-1}^\top)y_t$

Finally, we average the norms of the residuals on both views (with respect to their metric) leading to the performance measure $c_t = \frac{1}{2}(\frac{r_x^t{}^\top C_x^{t-1} r_x^t}{n_x} + \frac{r_y^t{}^\top C_y^{t-1} r_y^t}{n_y})$.

For the test session, we need to determine a threshold $\tau$ above which a change is detected. For a given subject, $\tau$ is defined as the minimum of the criterion $c$ evaluated over the change points in the three training sessions. The course of the reconstruction criterion $c$ over time is shown for each subject during each session in Figure 3. The vertical red dashed lines show the time step when the subject is asked to switch from a mental task to another. For every subject, the horizontal black dotted lines show the threshold $\tau$.

We observe in Figure 3 that the criterion $c_t$ rapidly rises at the exact moment of the change and that the criterion oscillates in the next few samples after a change. So we decided to adopt the following decision rule for the test data : if the criterion is greater than the threshold $\tau$ and if it was not previously (within 5 samples = 312ms) greater than the threshold $\tau$, a change is detected. Obviously, this is a naive rule and it could be improved by using more complex hypothesis test on the distribution of the criterion. However, as we want to highlight the effect of the CCA, this simple thresholding heuristic is clearly sufficient as made clear by Table 1.

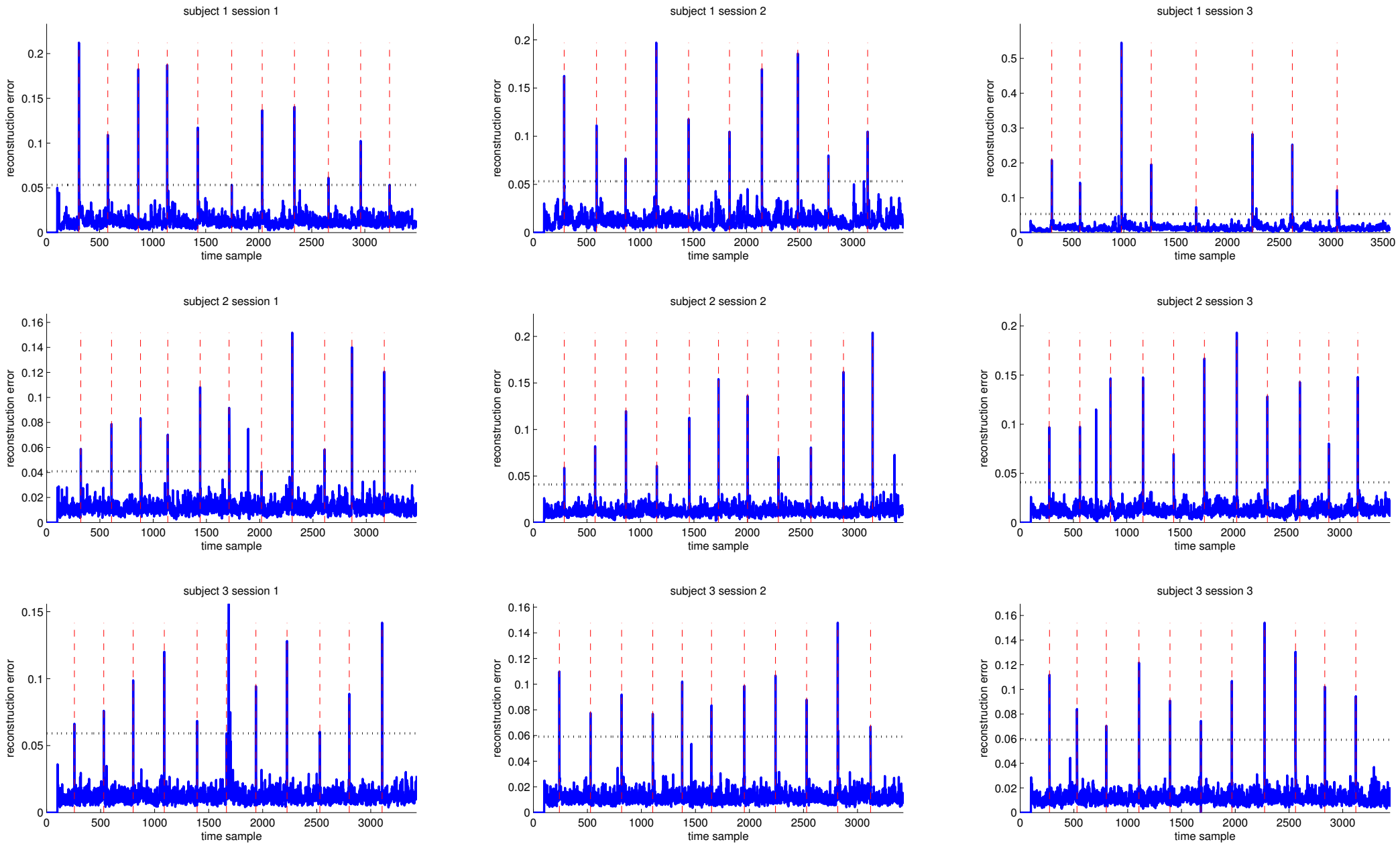

*Figure 3.* Application of the adaptive CCA to the training data of the subjects.

In order to evaluate the performance of our adaptive CCA method, we compare the AUC of the adaptive CCA to a state-of-the-art change detection method (Desobry et al., 2005). This method referred to as *KCD* in Table 1 is applied on the 96 precomputed PSD features with 100 samples sliding windows. The parameters ($\sigma$ of the Gaussian kernel, $\nu$ of the One-class SVM) algorithm were validated on the three training sessions in order to give the best AUC. As a comparison, we show the evolution of the KCD criterion in Figure 4. The CCA algorithm surprisingly outperforms the KCD algorithm. However, the AUC results may be explained by the strong *a priori* knowledge included in this approach by studying the correlation EEG signals between left and right hemispheres. Indeed this simple algorithm applied to features adapted to the problem outperforms a complex method applied on out of the box features.

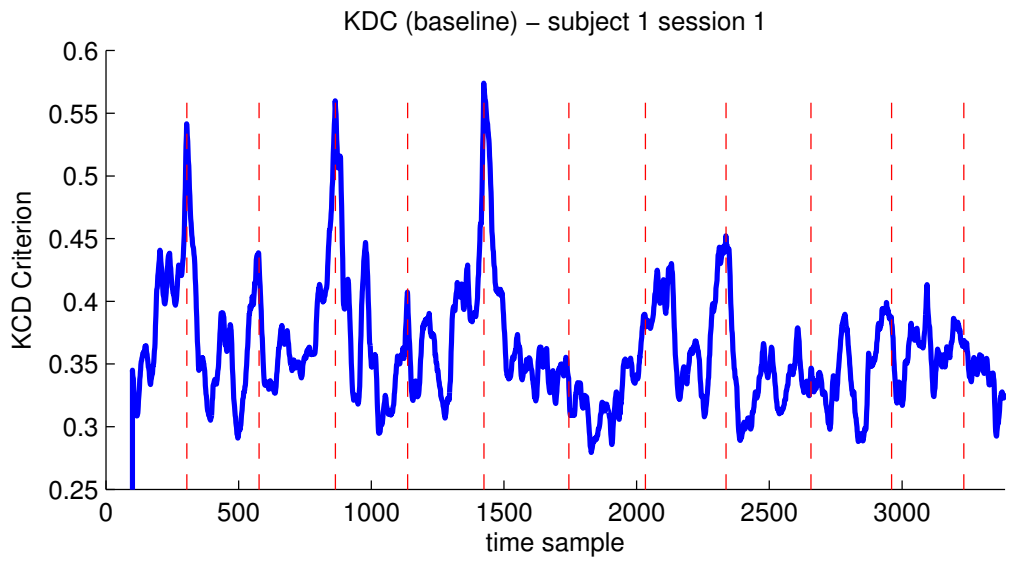

*Figure 4.* KCD criterion on the first session for subject 1 .

Our simple heuristic on the residuals of the adaptive CCA gives interesting results. It is notably able to detect changes without any delay and may be a good feature extraction for more robust change detection algorithm. Moreover, sensor and feature selection could also enhance the obtained results. Finally, using some stronger *a priori* on BCI data, this adaptive CCA approach could be extended to classify the task after the change is detected.

## 6. Conclusion and perspectives

We proposed an adaptive formulation of the classical CCA algorithm based on matrix manifolds. The Riemannian framework enabled us to build a fast and adaptive two-steps gradient algorithm. Moreover, we proposed an approach for change detection in the correlation of two time series. This simple approach was tested on a BCI dataset and gave promising results for detecting changes in mental tasks.

*Table 1.* Performance of the CCA for change detection on the BCI competition test dataset.

| | Subject 1 | Subject 2 | Subject 3 |
|---|---|---|---|
| threshold | 0.0533 | 0.0409 | 0.0591 |
| True Positives | 7 | 10 | 8 |
| False Positives | 3 | 0 | 1 |
| False Negatives | 1 | 1 | 3 |
| True Negatives | 3493 | 3461 | 3476 |
| #samples | 3504 | 3472 | 3488 |
| AUC (CCA) | 100 | 100 | 99.99 |
| AUC (KCD) | 81.12 | 68.63 | 58.35 |

We are currently exploring several extensions of this work. Our algorithm is based on a simple gradient and using the same manifold framework, a Newton method could be derived and should improve the distance to the batch solution. This paper focused on an adaptive formulation of the problem, its extension in an online learning framework (Warmuth & Kuzmin, 2006) should provide regret bounds for both views. Finally, Approximate Joint Singular Value Decomposition (AJSVD) (Congedo et al., 2011) is an extension of CCA handling the correlations between multiple views. To our knowledge, it has not been studied in an adaptive settings which would be of interest to BCI application.